%% file: main-TMM.tex
\documentclass[journal]{IEEEtran}
\PassOptionsToPackage{table}{xcolor}
\usepackage{xcolor}
\usepackage{amsmath,amsfonts}
\usepackage{array}
\usepackage[caption=false,font=normalsize,labelfont=sf,textfont=sf]{subfig}
\usepackage{textcomp}
\usepackage{stfloats}
\usepackage{url}
\usepackage{verbatim}
\usepackage{graphicx}
\usepackage{hyperref}
\hypersetup{hypertexnames=false}

\definecolor{blue}{RGB}{33,94,153}
\definecolor{red}{RGB}{192,0,0}
\definecolor{green}{RGB}{59,125,35}
\definecolor{orange}{RGB}{242,170,132}
\definecolor{purple}{RGB}{120,32,110}
\usepackage{multirow}
\usepackage{makecell}
\usepackage{arydshln}
\usepackage{float}

\usepackage{siunitx}  
\usepackage{booktabs}  
\usepackage{bm}

\def\BibTeX{{\rm B\kern-.05em{\sc i\kern-.025em b}\kern-.08em
    T\kern-.1667em\lower.7ex\hbox{E}\kern-.125emX}}
\usepackage{balance}
\usepackage{capt-of}
\begin{document}

\title{EventVL: Understand Event Streams via Multimodal \\ Large Language Model}
\author{Pengteng Li, Wuyang Li, Pinhao Song, Huizai Yao, Yunfan Lu$^*$, and Hui Xiong$^*$,~\IEEEmembership{Fellow,~IEEE}

\thanks{\textit{$^*$Corresponding authors: Yunfan Lu, Hui Xiong.}}
\thanks{Pengteng Li, Yunfan Lu, Huizai Yao and Hui Xiong are with the Thust of the Artificial Intelligence, The Hong Kong
University of Science and Technology (Guangzhou), Guangzhou 511453, China (email: \{pli807, ylu066, hyao032\}@connect.hkust-gz.edu.cn), {xionghui@ust.hk})}

\thanks{Pinhao Song is with the KU Leuven (email: \url{pinhao.song@kuleuven.be}).}

\thanks{Wuyang Li is with the École Polytechnique Fédérale de Lausanne (email: \url{wuyang.li@epfl.ch}).}

\thanks{This work was supported in part by the National Key R\&D Program of China (Grant No. 2023YFF0725001); in part by the National Natural Science Foundation of China (Grant No. 92370204); in part by the Guangdong Basic and Applied Basic Research Foundation (Grant No. 2023B1515120057); and in part by the Key-Area Special Project of Guangdong Provincial Ordinary Universities (Grant No. 2024ZDZX1007).}
}


\maketitle

\input{sec/0_abstract}

\input{sec/1_intro_V3_pinhao_wuyang}
\input{sec/2_related_work}

\input{sec/3_method_V2_pinhao}

\input{sec/4_experiments}

\input{sec/5_conclusion}

{   
    \clearpage
    \small
    \bibliographystyle{ieeenat_fullname}
    \bibliography{main}
}

\end{document}

%% file: sec/0_abstract.tex
\begin{abstract}
The event-based Vision-Language Model (VLM) recently has made good progress for practical vision tasks. However, most of these works just utilize CLIP for focusing on traditional perception tasks, which obstruct model understanding explicitly the sufficient semantics and context from event streams.
To address the deficiency, we propose \textbf{\textsc{EventVL}}, the novel generative event-based Multimodal  Large Language Model (MLLM) framework for explicit semantic understanding.
Specifically, to bridge the data gap for connecting different modalities semantics, we first annotate a large event-image/video-text dataset, containing \textbf{almost 1.4 million} high-quality pairs of data, which enables effective learning across various scenes, \textit{e.g.},  drive scene or human motion.
After that, we design Event Spatiotemporal Representation to fully explore the comprehensive information by diversely aggregating and segmenting the event stream. 
To further promote a compact semantic space, Dynamic Semantic Alignment is introduced to improve and complete sparse semantic spaces of events. Extensive experiments show that our \textsc{EventVL} can significantly surpass existing MLLM baselines in event captioning and question-answering tasks. We hope our research could contribute to the development of the event community. 
\end{abstract}


%% file: sec/1_intro_V3_pinhao_wuyang.tex
\section{Introduction}
\label{sec:intro}

Event cameras are sensors that asynchronously measure changes in intensity at each pixel with microsecond-level precision.
Compared to traditional cameras, event cameras offer several significant advantages~\cite{survey}, including a high dynamic range ($>120$ dB), resistance to motion blur, high temporal resolution, and extremely low power consumption.
These advantages have led to significant successes of event-based methods in different domains, such as robotics~\cite{he2024microsaccade} and imaging applications~\cite{LEOD,sun2022ess}. 
Even though event cameras offer richer semantic information—including continuous temporal motion and higher dynamic range lighting, exploration into high-level and open-world understanding tasks remains limited.

As VLMs~\cite{li2023blip,Grounding-DINO} achieve great successes in image understanding, the event-based community began to develop event-based VLMs for a comprehensive understanding of event data, which can be beneficial to downstream tasks~\cite{eventbind,EZSR} like open-vocabulary segments~\cite{OpenESS} for large-scale scene understanding. Most of these works focus on image-event-text pair alignment based on CLIP~\cite{CLIP} by bridging the modality gap between these data according to task scenarios.
Though successful, these CLIP-based methods limit themselves in traditional perception task~\cite{ExACT, eventclip} due to a coarse understanding of event data, which makes it hard to generalize to a fine-grained dynamic semantic task such as the detailed description for object motion pattern and the various relation between objects. It leads to the bottleneck in spatial understanding of many scenarios, for instance, autonomous driving or navigation. Recently, current multi-modal large language models (MLLMs)~\cite{llava,Gpt-4} have utilized images or videos as inputs to obtain an accurate and fine-grained text description, lifting visual perception to natural-language-based comprehension of the world. 
From our perspective, MLLMs have the potential to overcome the scene understanding limitations of event data and offer a versatile language interface for user interaction.

However, to train an event-based MLLM, we are faced with two main challenges.
First, a lack of high-quality text annotation in current image-event-text pair datasets hinders the model from capturing fine-grained semantic information~\cite{N-imagenet, hardvs}. In those datasets, the coarse text descriptions like \textcolor{gray}{``This is a \{category\}''} limit the model to the learning of category shapes and hinder the model from excavating more diverse event-based category knowledge such as colour or materials, which prohibits the fine-grained semantic understanding of event data. 
Second, different from images, event data, as a format of spatiotemporal point clouds, is incompatible with the current RGB encoder architecture. Moreover, preserving the highly spatiotemporal correlations within event data during feature extraction is challenging. An inadequate representation of events can impair event-image-text fine-grained alignment, resulting in sub-optimal scene understanding.


To solve these challenges mentioned above, we propose \textbf{\textsc{EventVL}}, an event-based MLLM framework, as shown in Figure~\ref{fig:framwork}. As an early exploration into generative event-based multimodal language modeling, \textsc{EventVL} is designed to address the unique properties of event streams—sparsity, asynchronous sampling, and missing surface attributes—through event-specific representation and alignment designs. Specifically, we propose \textbf{Event Spatiotemporal Representation (ESR)} to preserve the fine-grained spatiotemporal correlations within event data that traditional flat representations cannot retain, by hierarchically and adaptively decomposing event streams along both temporal and spatial dimensions. To further bridge the intrinsic event-image modality gap, we propose \textbf{Dynamic Semantic Alignment (DSA)}, which establishes the pretrained image latent space as a unified semantic anchor connecting events and language. By dynamically aligning event representations with this anchor space, DSA enables event features to inherit the rich attribute priors (e.g., texture, color, and material) encoded in large-scale pretrained image representations, thereby unlocking semantically grounded reasoning capabilities that are otherwise inaccessible to event-only modeling. As for data engineering, we annotate almost 1.4 million high-quality image-text-event paired data by utilizing current powerful open-source MLLM models~\cite{internvl-2.0}, which has surpassed the diverse MLLMs in many understanding benchmarks.
Comprehensive experiments verify that our proposed \textsc{EventVL} surpasses other SOTAs in zero/few-shot event captioning, event stream description generation tasks, and event-based dialogue. 
Furthermore, with a tiny number of parameters (\textbf{almost 2.3B}) compared to other MLLMs~\cite{Deepseek-vl,bai2025qwen3}, our model enables a low-cost for deployment to the real world. 
In summary, our contribution has three folds:

\begin{itemize}
    \item We present \textit{the novel Event-based MLLM framework}, named \textsc{EventVL}, which aligns the large-scale event encoder with the image encoder and LLM. The model demonstrates strong performance on event-based generative tasks such as description or dialogue. 
    \item \textit{Event Spatiotemporal Representation} is proposed for feature adaptive and efficient aggregation. We also propose a \textit{Dynamic Semantic Alignment} module for fine-grained feature alignment, resulting in a precise and comprehensive event-based understanding. 
    \item We propose the high quality event-image-text paired datasets, which contain \textbf{almost 1.4 million} paired data across various domains. As we known, it is the biggest multi-modalities pair dataset in the event community.
\end{itemize}




%% file: sec/2_related_work.tex
\section{Related Work}
\label{sec:Related work}

\noindent \textbf{Open-Source MLLMs.} Open-source MLLMs align pretrained vision encoders with LLMs for multimodal reasoning. BLIP-2~\cite{li2023blip} uses a lightweight Q-Former to bridge frozen encoders and language models, while Flamingo~\cite{Flamingo} injects visual features through gated cross-attention for few-shot reasoning. Later models~\cite{llava,Deepseek-vl,Qwen2-vl} combine visual instruction tuning with large-scale image-text pretraining. However, their frame-based inputs can be unreliable under motion blur or difficult lighting, motivating event streams as a complementary modality with high temporal resolution and dynamic range.

\noindent \textbf{Event-based VLMs.} Early work learned event representations through self-supervision~\cite{Maskedeventmodeling, Event-camera-data-pre-training}. CLIP-inspired methods align event and text by converting events to image-like representations~\cite{eventclip} or adding multimodal alignment modules~\cite{eventbind, CEIA}, supporting classification and segmentation~\cite{hardvs, OpenESS}. These approaches establish useful event-image-text correspondences, but remain largely discriminative and do not directly support event-conditioned language generation or multi-turn interaction.

\noindent \textbf{Event-based MLLMs.} EventGPT~\cite{liu2025eventgpt}, Llafea~\cite{Llafea}, EventSTU~\cite{EventSTU}, and Talk2Event~\cite{talk2event} explore generative event understanding through frame-event fusion, event-guided video reasoning, or object grounding. These methods demonstrate the potential of language interfaces for event data, but may rely on additional visual modalities or focus on a specific downstream task. EventVL instead targets event-only captioning, VQA, and dialogue by addressing the event-image gap with Event Spatiotemporal Representation and Dynamic Semantic Alignment.


\begin{figure*}[t]
    \centering
    \includegraphics[width=\linewidth]{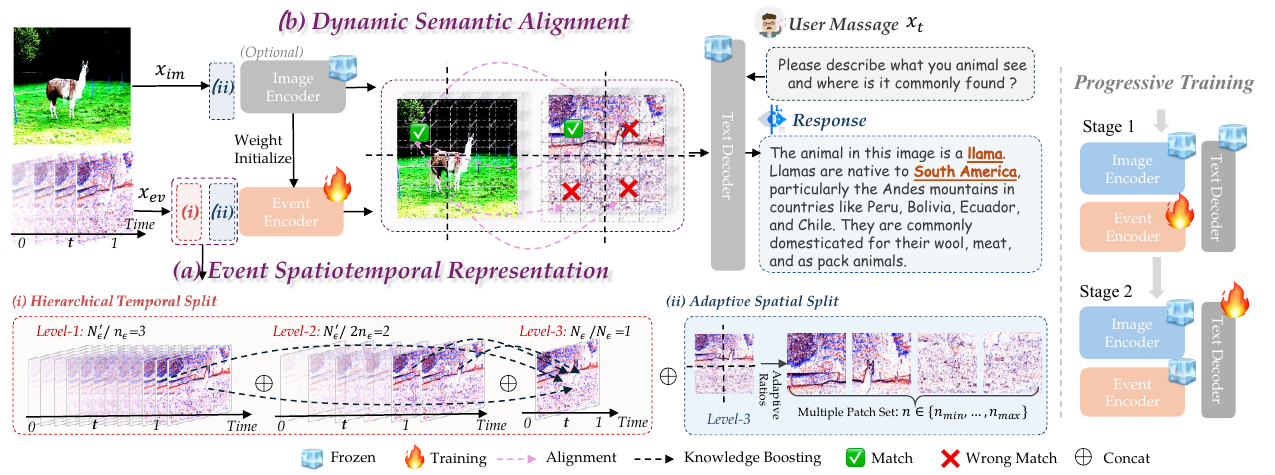}
    \vspace{-0.4cm}
    \caption{Overview of our proposed \textsc{EventVL}. \textsc{EventVL} mainly consists of  two parts: \textit{\textcolor{purple}{(a) Event Spatiotemporal Representation}} and \textit{\textcolor{purple}{(b) Dynamic Semantic Alignment.}} First, we introduce Event Spatiotemporal Representation for extracting diverse event information comprehensively, which consists of two split techniques --\textit{\textcolor{red!70}{(i) Hierarchical Temporal Split}} and \textit{\textcolor{blue!70}{(ii) Adaptive Spatial Split.}} Then, we utilize Dynamic Semantic Alignment to further push the image and event feature space in fine-grained alignment. Finally, we perform Progressive Training strategy to make \textsc{EventVL} better.}
    
    \label{fig:framwork}
    \vspace{-0.3cm}
\end{figure*}

%% file: sec/3_method_V2_pinhao.tex
\section{Method}
\label{sec:Method}

The overall workflow of \textsc{EventVL} is as shown in Figure~\ref{fig:framwork}, which is composed of a frozen image encoder, a trainable event encoder, and a frozen text decoder. 
Given batch-wise image-event-text pair data $\{(x_{im},x_{ev},\{x^l_{t}\}_{l=1}^{L})\}$, where event streams $x_{ev}$ contains $N_\epsilon$ event points, $x_{im}$ and $\{x^l_{t}\}_{l=1}^{L}$ denotes the image and text token list.
We introduce Event Spatiotemporal Representation containing fine-grained spatiotemporal event semantic information, which is obtained by two split techniques -- Hierarchical Temporal Split and Adaptive Spatial Split. 
After that, Dynamic Semantic Alignment is proposed for further projecting event data on image feature space in a compact latent space. We use the red-blue color map as the event representation consistent with recent approaches~\cite{CEIA,OpenESS}. This choice minimizes the gap between the event representation and the natural images for pre-trained models, thereby simplifying cross-modal alignment.


\subsection{Event Spatiotemporal Representation}
\label{sec:ESR}
Event streams carry rich information about the captured scene~\cite{wang2021joint}. To enable effective scene understanding, it is crucial to develop a comprehensive representation of event streams that aligns with neural networks for encoding spatiotemporal semantics.
Hence, we propose Event Spatiotemporal Representation which divides and captures fine-grained temporal and spatial information. Hierarchical Temporal Split and Adaptive Spatial Split are proposed to obtain different groups of event frames containing temporal and spatial information, respectively. Finally, all event frames are concatenated, gathering the fine-grained spatiotemporal information.


\label{sec:ESR/HTS}
\noindent \textbf{Hierarchical Temporal Split.}  Due to the temporal dynamics in event data, there is a critical need for a novel approach to extract rich temporal semantics. Our method diverges from prior studies that have opted for introducing a multitude of parameters~\cite{eventbind}. Instead, we propose Hierarchical Temporal Split to aggregate event temporal information in various levels, without the need for any additional parameters. As shown in Figure~\ref{fig:framwork} (i), given an event stream $x_{ev}$ containing $N_\epsilon$ event points, we firstly define the event points per frame $\lambda n_\epsilon, \lambda \in \{1,2\}$ for splitting the overall event stream into several groups, and transform these groups into RGB-style representations. After splitting hierarchically, we expand the events as multi-level events $\hat{\bm{x}}_{ev} = \{\{\hat{x}^r_{ev1}\}^{N_1}_{r=1}, \{\hat{x}^r_{ev2}\}^{N_2}_{r=1}, \hat{x}_{ev3}\}$ with three levels contains $N_1$, $N_2$, and $1$ RGB frames, respectively, where $N_1 = \frac{N_\epsilon}{n_\epsilon}$, $N_2 = \frac{N_\epsilon}{2n_\epsilon}$, and $\hat{x}^r_{ev*} \in \mathbb{R}^{H \times W}$ denotes the transformed RGB-style representations with a height of $H$ and a width of $W$. The level-3 event representation $\hat{x}_{ev3}$ aggregates all event points. In implementation, we set the $N_\epsilon$ as the fixed number and use padding for event streams containing less than $N_\epsilon$ event points. Finally, these temporal splits are sent to Adaptive Spatial Split for further spatial semantics extraction. 

\label{sec:ESR/ASS}
\noindent \textbf{Adaptive Spatial Split.} Previous research has focused on encoding complete event representations and aligning them with corresponding images within the feature space. While this method has led to better performance~\cite{EZSR, CEIA}, it requires resizing event frames to a low resolution, such as $224 \times 224$. This resizing results in a significant loss of spatial information, particularly for high-resolution events like those in \textsc{N-ImageNet}~\cite{N-imagenet}, which can measure $480 \times 640$. This issue worsens under poor imaging conditions, such as overexposure or rolling shutter effects, leading to even greater information loss.
Hence, Adaptive Spatial Split is introduced to explore full spatial event-based category information, as shown in Figure~\ref{fig:framwork} (ii), which adaptively splits event representation or images into multiple patches uniformly in a high-resolution manner.

To preserve natural aspect ratios, we construct $\mathcal{K}$ by enumerating and sorting all integer tile ratios $(i,j)$ whose area satisfies $n_{\min}\leq i j\leq n_{\max}$; we set $n_{\min}=1$ and $n_{\max}=6$. For each input, we select the ratio with the smallest aspect-ratio error, breaking ties by choosing a layout whose area does not exceed twice the input area, and resize the resulting patches to the target dimensions.
In detailed, given the multi-level events $\hat{\bm{x}}_{ev} = \{\{\hat{x}^r_{ev1}\}^{N_1}_{r=1}, \{\hat{x}^r_{ev2}\}^{N_2}_{r=1}, \hat{x}_{ev3}\}$, we only apply Adaptive Spatial Split to the level-3 event frame $\hat{x}_{ev3}$ and obtain multiple event patches spatially $\{\{\hat{x}^r_{ev3}\}^{N_p}_{r=1}\}$ ($ N_p =6$). We concatenate multiple event patches with the origin multi-level events, obtaining Event Spatiotemporal Representation $\tilde{\bm{x}}_{ev} = \{\{\hat{x}^r_{ev1}\}^{N_1}_{r=1}, \{\hat{x}^r_{ev2}\}^{N_2}_{r=1}, \hat{x}_{ev3}, \{\hat{x}^r_{ev3}\}^{N_{p}}_{r=1} \}$.


\subsection{Dynamic Semantic Alignment}
In this section, we aim to align event and image within the same latent space. We begin by inputting the event representation $\tilde{\bm{x}}{ev}$ into the event encoder, which is composed of multiple attention blocks. This process yields event embeddings $\{\{\phi^r_{ev1}\}^{N_1}_{r=1}, \{\phi^r_{ev2}\}^{N_2}_{r=1}, \phi_{ev3}, \{\phi^r_{ev3}\}^{N_{p}}_{r=1} \}$ that are enriched with exchanged spatiotemporal knowledge and enhanced semantics. For feature alignment, we select $\bm{\phi}_{ev} = \{\phi_{ev3}, \{\phi^r_{ev3}\}^{N_{p}}_{r=1} \}$.
To obtain the corresponding image embeddings for alignment, we can perform the Adaptive Spatial Split for image data $x_{im}$ and get the segmented image patches concatenated with $x_{im}$ to obtain $\{ x_{im}, \{x^r_{im}\}^{N_p}_{r=1} \}$, which is inputted into the image encoder to obtain image embeddings $\bm{\phi}_{im} = \{ \phi_{im}, \{\phi^r_{im}\}^{N_p}_{r=1} \}$. Following the feature extraction by the attention blocks, each frame in $\bm{\phi}{ev}$ is sufficiently endowed with global temporal semantics and enhanced spatial knowledge. Finally, we conduct a cosine alignment between $\bm{\phi}_{ev}$ and $\bm{\phi}_{im}$.


Compared to ordinary images, event representation lacks detailed surface descriptions, such as material and texture. It motivates us to project events into the same compact feature space as images, thereby implicitly learning missing attribute descriptions. Hence, we leverage simple cosine similarity loss to maintain their mutual information,
\begin{equation} 
    \mathcal{L}_{c} = 1 - \mathop{\textnormal{cos}}(\bm{\phi}_{ev}, \bm{\phi}_{im}) = 1 - \frac{\bm{\phi}_{ev} \cdot \bm{\phi}_{im}}{\Vert \bm{\phi}_{ev} \Vert \Vert \bm{\phi}_{im} \Vert}.
\label{eq:cos}
\end{equation}
By applying this loss, spatiotemporal fine-grained alignment can be achieved, and event data are projected into the same latent space as image data.


\subsection{Framework Optimization}
Generally, maximizing the likelihood function below to align the representation spaces of image/video and text is a widely-used approach for pre-training~\cite{llava, xu2023u}. We also leverage this pattern to our event-text alignment. For a given event representation embeddings $\bm{\phi}_{ev}$ and a conversation list of $L$ text tokens $\bm{x}_t = \{x^1_t, x^2_t, ..., x^L_t\}$, the likelihood of this list can be written as follows: 
\begin{equation} 
\vspace{-0.1cm}
    p(\bm{x}_t \mid \bm{\phi}_{ev}, \bm{x}_{\textnormal{instruct}}) = \prod_{l=1}^{L}\phi_t(x_{t}^{l} \mid\bm{\phi}_{ev}, \bm{x}_{\textnormal{instruct}}, x_{t}^{1:l-1}),
\end{equation}
where $\bm{x}_{\textnormal{instruct}}$ is the instruction token, and $\phi_t(\cdot)$ is the text decoder. Then we perform event-text alignment by minimizing the negative log-likelihood, as follows:
\begin{equation} \label{ce}
    \mathcal{L}_{ev, t} = - \mathop{\textnormal{log}} p(\bm{x}_t \mid \bm{\phi}_{ev}, \bm{x}_{\textnormal{instruct}}).
\end{equation}
We also have observed 
that text inputs often include detailed descriptions such as texture and color, which are not present in the corresponding event data due to the nature of event imaging. Relying solely on event-text alignment can lead to suboptimal outcomes, as it may focus on missing semantic elements. Therefore, to refine the embedding alignment, we incorporate the image embedding as a prior in Equation~\ref{ce}, effectively limiting the search space and enhancing the overall alignment process:
\begin{equation} 
    \mathcal{L}_{ev,im, t} = - \mathop{\textnormal{log}} p(\bm{x}_t \mid \frac{1}{2}(\bm{\phi}_{ev}+\bm{\phi}_{im}), \bm{x}_{\textnormal{instruct}}).
\end{equation}
The overall training objectives can be written as follows:
\begin{equation}
    \mathcal{L} = \frac{1}{2} \lambda_1  (\mathcal{L}_{ev, t} + \mathcal{L}_{ev, im, t}) + \lambda_2 \mathcal{L}_c.
    \label{opm}
\end{equation}
Note that the loss is computed in a batch and we omit the batch notation for clarity.

\subsection{Progressive Training and Inference}
We adopt a two-stage approach for model training. In the first stage, we fine-tune the event encoder while keeping all other modules frozen, utilizing the \textsc{EventVL-Base} dataset. This step is designed to enhance the foundational event comprehension ability of the image-based encoder. In the second stage, we fine-tune the text decoder while freezing the other modules, using the \textsc{EventVL-QA} dataset to improve the model's fine-grained analytical capabilities for event-text pairs. This progressive training strategy enables us to develop a world model that comprehensively understands events across various scenarios. Furthermore, \textsc{EventVL} supports two inference pipelines. The first pipeline processes only the event modality as input, while the second integrates both image and event modalities. Finally, event embeddings are directly combined with image embeddings to form the input representation, obtaining more comprehensive understanding of the event.









\begin{table}[t] 
\centering
\caption{Statistical information of the datasets used in \textsc{EventVL}. 
``Caption'' denotes the \textsc{EventVL} dataset with descriptive annotations, 
while ``QA'' denotes the proposed \textsc{EventVL-QA} dataset with question-answer pairs.}
\resizebox{\columnwidth}{!}{
\begin{tabular}{lccccrr}
\toprule
\textbf{Task} & \textbf{Dataset} & \textbf{Type} & \textbf{Resolution} & \textbf{Split} & \textbf{Category} & \textbf{Scale} \\
\midrule

\multirow{7}{*}{Caption} 
& \multirow{2}{*}{\textsc{N-ImageNet}~\cite{N-imagenet}} & \multirow{2}{*}{Image} & \multirow{2}{*}{$480 \times 640$} 
& train & 800 & 1,023,907 \\
& & & & val & 200 & 257,260 \\

& \multirow{2}{*}{\textsc{HARDVS}~\cite{hardvs}} & \multirow{2}{*}{Video} & \multirow{2}{*}{$346 \times 260$} 
& train & 240 & 61,349 \\
& & & & val & 60 & 13,915 \\

& \textsc{DSEC}~\cite{Dsec} & Video & $640 \times 480$ & train & 42 & 2,173 \\

& \multirow{2}{*}{\textsc{N-Caltech101}~\cite{n-caltech}} & \multirow{2}{*}{Image} & \multirow{2}{*}{$302 \times 245$} 
& train & 80 & 29,550 \\
& & & & val & 21 & 5,290 \\

\midrule

\multirow{5}{*}{QA} 
& \multirow{2}{*}{\textsc{N-ImageNet}~\cite{N-imagenet}} & \multirow{2}{*}{Image} & \multirow{2}{*}{$480 \times 640$} 
& train & 160 & 16,000 \\
& & & & val & 40 & 4,000 \\

& \multirow{2}{*}{\textsc{HARDVS}~\cite{hardvs}} & \multirow{2}{*}{Video} & \multirow{2}{*}{$346 \times 260$} 
& train & 480 & 9,600 \\
& & & & val & 120 & 2,400 \\

& \textsc{DSEC}~\cite{Dsec} & Video & $640 \times 480$ & train & 42 & 43,460 \\

\midrule
\cellcolor{blue!20}\textit{\textbf{Total (Caption)}} & \multicolumn{4}{c}{\cellcolor{blue!20}-} & \cellcolor{blue!20}\textbf{1,443} & \cellcolor{blue!20}\textbf{1,393,444} \\
\cellcolor{blue!20}\textit{\textbf{Total (QA)}} & \multicolumn{4}{c}{\cellcolor{blue!20}-} & \cellcolor{blue!20}\textbf{842} & \cellcolor{blue!20}\textbf{75,460} \\

\bottomrule
\end{tabular}
}
\label{tab:eventvl_datasets}
\end{table}

\section{Data Engineering}
\label{sec:DataEngineering}

A high-quality dataset is crucial for training an MLLM, yet existing image/video-event datasets~\cite{N-imagenet, hardvs} mainly provide coarse labels such as \textcolor{gray}{"This is a {category}"}. Such annotations lack fine-grained attributes and may introduce irrelevant information; Table~\ref{tab:dataset performance} confirms that directly training on them can reduce performance. We therefore design a unified generation pipeline for static images, human motion, and driving scenes, using \textsc{InternVL2-76B}~\cite{internvl-2.0} to produce detailed descriptions and dialogue data at lower cost than commercial engines.

\noindent \textbf{Coarse Generation.} We manually define dataset-specific prompts and randomly sample one per image-event pair before sending the image and event frames to \textsc{InternVL2-76B}. For video-event datasets, using every frame is inefficient~\cite{OpenESS}; we therefore uniformly sample 14 frames (or all frames when fewer are available) to capture long-term information at manageable cost.

\noindent \textbf{Manual Check.} To avoid checking every pair, we sample descriptions by category or scene (1,000, 101, 300, and 20 groups for \textsc{N-ImageNet}, \textsc{N-Caltech101}, \textsc{HARDVS}, and \textsc{DSEC}). Unsatisfactory groups are regenerated with modified prompts; remaining errors are manually reviewed and corrected.

\begin{table}[t] 
\centering
\caption{Exploration experiments of using the untraining \textsc{EventVL} initialized from \textsc{InternVL2-2B}~\cite{internvl-2.0}. ``wo ev./im.'' denotes that we don't use the event or image modality as input.}
\resizebox{0.9\columnwidth}{!}{
\begin{tabular}{c|c|cccc}
\toprule
\textbf{Dataset} & \textbf{Model} & \textbf{Bleu-4} & \textbf{METEOR} & \textbf{ROUGE-L} & \textbf{CIDEr} \\
\midrule

\multirow{3}{*}{\textsc{N-ImageNet}} & - & 0.179 & 0.203 & 0.437 & 1.306 \\
& w/o ev. & \textbf{0.218} & \textbf{0.233} & \textbf{0.471} & \textbf{1.572} \\
&  w/o im. & 0.052 & 0.098 & 0.252 & 0.203 \\
\midrule

\multirow{3}{*}{\textsc{N-Caltech101}} & - & 0.313 & 0.255 & 0.562 & 0.874 \\
& w/o ev.& \textbf{0.362} & \textbf{0.297} & \textbf{0.614} & \textbf{1.123} \\
&  w/o im.& 0.083 & 0.130 & 0.326 &  0.111 \\
\midrule

\multirow{3}{*}{\textsc{HARDVS}} & - & 0.160 & 0.218 & 0.422 & 0.714 \\
& w/o ev. & \textbf{0.227} & \textbf{0.242} & \textbf{0.499} & \textbf{1.219} \\
&  w/o im. & 0.012 & 0.099 & 0.189 & 0.040 \\

\bottomrule

\end{tabular}
}
\label{tab:dataset performance}
\end{table}

Finally, we construct \textsc{EventVL-Base} with 1,393,444 descriptive multimodal pairs and \textsc{EventVL-QA} with 75,460 QA pairs. We define 10, 20, and 20 sequential questions per category for \textsc{N-ImageNet}, \textsc{HARDVS}, and \textsc{DSEC}, respectively. Their statistics are summarized in Table~\ref{tab:eventvl_datasets}.


%% file: sec/4_experiments.tex
\section{Experiments}
\label{sec:Experiments}


\begin{table*}[t] 
\centering
\caption{Quantitative results of zero-shot event-based caption generation on different datasets using only event inputs. \textsc{EventVL} achieves competitive performance compared with state-of-the-art MLLMs. \colorbox{gray!20}{Gray color} denotes video-based MLLMs, and \colorbox{blue!5}{\textsc{InternVL-E}} denotes \textsc{InternVL} finetuned on our proposed datasets.}
\resizebox{0.9\linewidth}{!}{
\begin{tabular}{l|c|c|ccccccc}
\toprule
\textbf{Dataset} & \textbf{Model} & \textbf{\#param} & \textbf{Bleu-1} & \textbf{Bleu-2} & \textbf{Bleu-3} & \textbf{Bleu-4} & \textbf{METEOR} & \textbf{ROUGE-L} & \textbf{CIDEr} \\
\midrule

\multirow{7}{*}{\textsc{N-ImageNet}~\cite{N-imagenet}} & \textsc{Qwen2-VL-Chat}~\cite{bai2025qwen3} & 10B & 0.346 & 0.153 & 0.069 & 0.033 & 0.096 & 0.287 & 0.092\\
& \textsc{DeepSeek-VL-Chat}~\cite{Deepseek-vl} & 7B & 0.217 & 0.096 & 0.042 & 0.022 & 0.113 & 0.233 & 0.034\\
& \textsc{LLaVA-v1.6}~\cite{llava} & 7B & 0.294 & 0.148 & 0.073 & 0.039 & 0.105 & 0.269 & 0.098\\
& \cellcolor{gray!20}\textsc{Video-LLaVA}~\cite{Video-llava} & \cellcolor{gray!20}7B & \cellcolor{gray!20}0.368 & \cellcolor{gray!20}0.183 & \cellcolor{gray!20}0.074& \cellcolor{gray!20}0.047 & \cellcolor{gray!20}0.118& \cellcolor{gray!20}0.298& \cellcolor{gray!20}0.134\\
& \cellcolor{gray!20}\textsc{VideoLLaMA2}~\cite{damonlpsg2024videollama2} & \cellcolor{gray!20}7B & \cellcolor{gray!20}\textbf{0.393} & \cellcolor{gray!20}0.200 & \cellcolor{gray!20}0.107 & \cellcolor{gray!20}0.061 & \cellcolor{gray!20}0.121 & \cellcolor{gray!20}0.326 & \cellcolor{gray!20}0.150 \\
& \cellcolor{blue!5}\textsc{InternVL-E}~\cite{llava} & \cellcolor{blue!5}2.3B & \cellcolor{blue!5} 0.360 & \cellcolor{blue!5} 0.233 & \cellcolor{blue!5} 0.144 & \cellcolor{blue!5} 0.092 & \cellcolor{blue!5} 0.138 & \cellcolor{blue!5} 0.354 & \cellcolor{blue!5} 0.423 \\
& \cellcolor{blue!20} \textbf{\textsc{EventVL (Ours)}} & \cellcolor{blue!20}2.3B & \cellcolor{blue!20}0.367 & \cellcolor{blue!20}\textbf{0.248} & \cellcolor{blue!20}\textbf{0.183} & \cellcolor{blue!20}\textbf{0.125} & \cellcolor{blue!20}\textbf{0.157} & \cellcolor{blue!20}\textbf{0.397} & \cellcolor{blue!20}\textbf{0.853} \\
\midrule

\multirow{7}{*}{\textsc{HARDVS}~\cite{hardvs}} & \textsc{Qwen2-VL-Chat}~\cite{bai2025qwen3} & 10B & 0.214 & 0.091 & 0.042 & 0.023 & 0.097 & 0.214 & 0.066\\
& \textsc{DeepSeek-VL-Chat}~\cite{Deepseek-vl} & 7B & 0.133 & 0.046 & 0.015 & 0.006 & 0.100 & 0.170 & 0.021\\
& \textsc{LLaVA-v1.6}~\cite{llava} & 7B & 0.196 & 0.087 & 0.032 & 0.021 & 0.097 & 0.202 & 0.054\\
& \cellcolor{gray!20}\textsc{Video-LLaVA}~\cite{Video-llava} & \cellcolor{gray!20}7B & \cellcolor{gray!20}0.299 & \cellcolor{gray!20}0.144 & \cellcolor{gray!20}0.087 & \cellcolor{gray!20}0.092 & \cellcolor{gray!20}0.103 & \cellcolor{gray!20}0.257 & \cellcolor{gray!20}0.124 \\
& \cellcolor{gray!20}\textsc{VideoLLaMA2}~\cite{damonlpsg2024videollama2} & \cellcolor{gray!20}7B & \cellcolor{gray!20}0.336 & \cellcolor{gray!20}0.186 & \cellcolor{gray!20}0.119 & \cellcolor{gray!20}0.073 & \cellcolor{gray!20}0.134 & \cellcolor{gray!20}0.348 & \cellcolor{gray!20}0.146 \\

& \cellcolor{blue!5}  InternVL-E~\cite{internvl-2.0} & \cellcolor{blue!5}2.3B & \cellcolor{blue!5} 0.457 & \cellcolor{blue!5} 0.322 & \cellcolor{blue!5} 0.289  & \cellcolor{blue!5} 0.245 & \cellcolor{blue!5} 0.211 & \cellcolor{blue!5} 0.498 & \cellcolor{blue!5} 1.845 \\
&  \cellcolor{blue!20} \textbf{\textsc{EventVL (Ours)}} & \cellcolor{blue!20}2.3B & \cellcolor{blue!20}\textbf{0.623} & \cellcolor{blue!20}\textbf{0.534}  & \cellcolor{blue!20}\textbf{0.464} & \cellcolor{blue!20}\textbf{0.407} & \cellcolor{blue!20}\textbf{0.326} & \cellcolor{blue!20}\textbf{0.637} & \cellcolor{blue!20}\textbf{2.619} \\
\midrule

\multirow{7}{*}{\textsc{N-Caltech101}~\cite{n-caltech}} & \textsc{Qwen2-VL-Chat}~\cite{Qwen2-vl} & 10B & 0.197 & 0.087 & 0.040 & 0.021 & 0.067 & 0.208 & 0.152\\
& \textsc{DeepSeek-VL-Chat}~\cite{Deepseek-vl} & 7B & 0.161 & 0.071 & 0.033 & 0.017 & 0.092 & 0.191 & 0.100\\
& \textsc{LLaVA-v1.6}~\cite{llava} & 7B & 0.240 & 0.119 & 0.061 & 0.034 & 0.083 & 0.203 & 0.186\\
& \cellcolor{gray!20}\textsc{Video-LLaVA}~\cite{Video-llava} & \cellcolor{gray!20}7B & \cellcolor{gray!20}0.246 & \cellcolor{gray!20}0.133 & \cellcolor{gray!20}0.082 & \cellcolor{gray!20}0.041 & \cellcolor{gray!20}0.095 & \cellcolor{gray!20}0.224 & \cellcolor{gray!20}0.255\\
& \cellcolor{gray!20}\textsc{VideoLLaMA2}~\cite{damonlpsg2024videollama2} & \cellcolor{gray!20}7B & \cellcolor{gray!20}0.319 & \cellcolor{gray!20}0.167 & \cellcolor{gray!20}0.094 & \cellcolor{gray!20}0.057 & \cellcolor{gray!20}0.106 & \cellcolor{gray!20}0.273 & \cellcolor{gray!20}0.358 \\
& \cellcolor{blue!5}\textsc{InternVL-E}~\cite{llava} & \cellcolor{blue!5}2.3B & \cellcolor{blue!5} 0.421 & \cellcolor{blue!5} 0.298 & \cellcolor{blue!5} 0.163 & \cellcolor{blue!5} 0.144 & \cellcolor{blue!5} 0.328 & \cellcolor{blue!5} 0.345 & \cellcolor{blue!5} 0.411 \\
& \cellcolor{blue!20} \textbf{\textsc{EventVL (Ours)}} & \cellcolor{blue!20}2.3B & \cellcolor{blue!20}\textbf{0.574} & \cellcolor{blue!20}\textbf{0.408} & \cellcolor{blue!20}\textbf{0.283} & \cellcolor{blue!20}\textbf{0.223} & \cellcolor{blue!20}\textbf{0.183} & \cellcolor{blue!20}\textbf{0.477} & \cellcolor{blue!20}\textbf{0.599} \\
\bottomrule

\end{tabular}
}


\label{tab:zsdg}
\end{table*}

\begin{figure*}[t]
    \centering    \includegraphics[width=\linewidth]{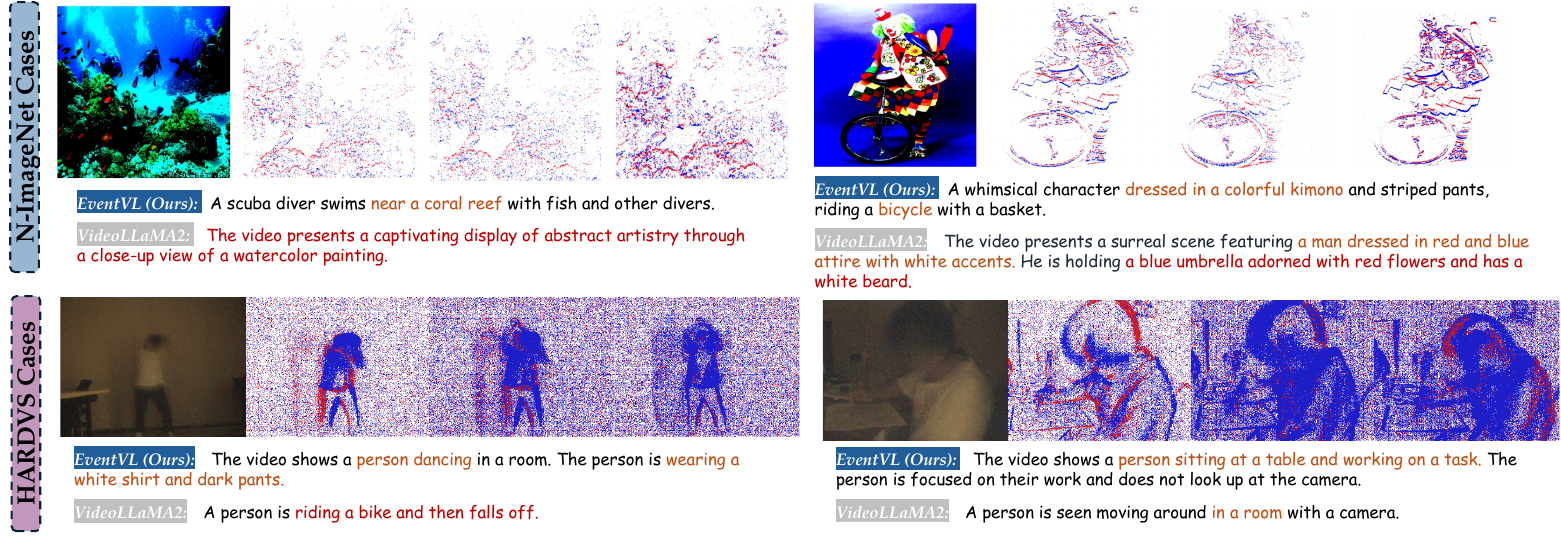}
    \vspace{-0.4cm}
    \caption{Qualitative results about zero-shot caption generation by utilizing \textsc{N-ImageNet}~\cite{N-imagenet} and \textsc{HARDVS}~\cite{hardvs} val set from Event-Base dataset. \textcolor{red}{Red color} denotes model wrong description and \textcolor{orange}{orange color} denotes the accurate object description. \underline{Underline text} means redundant description. It shows that \textsc{EventVL} can easily distinguish and analyze the vague objects imaging in the complex scenes like low light or motion-blur based on event streams.}
    \label{fig:illu1}
    \vspace{-0.2cm}
\end{figure*}

\subsection{Implementation}

\noindent \textbf{General Setting.} 
In Dynamic Semantic Alignment, we set $n_{\min}=1$, $n_{\max}=6$, and $\lambda_1=\lambda_2=1$. 
Due to the different event densities, the number of event points per frame is set to $n_\epsilon=40000$ for \textsc{N-ImageNet} and $n_\epsilon=20000$ for \textsc{N-Caltech101}. 
If the total event number $N_\epsilon$ is insufficient for the hierarchical representation in Section~\ref{sec:ESR/HTS}, the event stream is padded during preprocessing. 
For video-event datasets (\textsc{HARDVS} and \textsc{DSEC}) training, we directly apply cosine similarity (Eq.~\ref{eq:cos}) for modality alignment without using Event Spatiotemporal Representation, as these datasets already contain sufficient temporal information. 
To address dataset scale imbalance, we adopt weighted sampling (0.6 for \textsc{N-ImageNet}, 0.3 for \textsc{HARDVS}, and 0.1 for \textsc{DSEC}). 
Training is conducted on 8 NVIDIA GTX A6000 GPUs for 144 hours with batch size 2 using DeepSpeed ZeRO-1. Following prior event-based VLM works~\cite{eventbind,eventclip,CEIA}, event streams are converted into event frames and fed into MLLMs. 
To further evaluate event-based MLLMs, we adopt \textsc{InternVL}~\cite{internvl-2.0} and \textsc{Qwen3-VL}~\cite{bai2025qwen3} as base models. 
For both models, an event encoder is initialized from the pretrained image encoder and the models are finetuned on our proposed datasets for one epoch under identical settings, resulting in \textsc{InternVL-E} and \textsc{Qwen3-VL-E}.

\noindent \textbf{Fundamental Module.} We selected the InternViT-300M-448px as the foundational architecture for both the image and event encoders, due to its proven effectiveness across various image and video understanding benchmarks. For the text decoder, we chose \textsc{InternLM2-1.8B}, which has demonstrated significant improvements in reasoning and coding tasks.

\noindent \textbf{Zero-shot Setting.} We use the \textsc{N-ImageNet}, \textsc{HARDVS}, and \textsc{DSEC} train splits for training. To prevent overfitting, we perform training for 1 epoch. We evaluate \textsc{EventVL} on the validation sets for zero-shot description generation, using only event data as input.

\noindent \textbf{Few-shot Setting.} For the few-shot experiments, we sample $n$-shot data from the entire \textsc{N-ImageNet}, \textsc{HARDVS}, and \textsc{DSEC} datasets for training. The remaining datasets, including \textsc{N-Caltech101}, are used for evaluation. We perform all few-shot training for 3 epochs to ensure sufficient learning of event-based category knowledge.

\noindent \textbf{Evaluation Metrics.} Similar to previous works~\cite{hawk,bleu}, we use BLEU, METEOR, ROUGE-L, CIDEr, and VQA~\cite{Videogpt} to measure lexical accuracy, semantic similarity, and answer quality. For VQA, we use structured prompts and then leverage GPT for quality assessment on a 1--5 scale, establishing a comprehensive benchmark with the \textsc{EventVL-QA} datasets. This protocol enables a consistent evaluation of \textsc{EventVL} across generation and question-answering tasks.

\subsection{Datasets}

\noindent \textbf{Datasets.} We conduct experiments on several representative event benchmarks. 
\textsc{N-ImageNet}~\cite{N-imagenet} is the largest event-based dataset, containing nearly 1.2M event streams across 1,000 classes. 
\textsc{HARDVS}~\cite{hardvs} is a large-scale benchmark for event-based action recognition with 300 action categories and over 100K recordings. 
\textsc{DSEC}~\cite{Dsec} focuses on autonomous driving scenarios and provides complex real-world event streams. 
We further adopt \textsc{N-Caltech101}~\cite{n-caltech}, which contains 8,246 samples from 101 classes, for evaluating zero-shot and few-shot generalization.


\noindent \textbf{Processing Details.} 
In each annotation round, a problem is randomly sampled from the predefined task list. 
For \textsc{N-ImageNet}, we use the full event-pairs dataset for training. 
\textsc{N-Caltech101} is only used to evaluate the effectiveness of \textsc{EventVL}. 
For \textsc{HARDVS}, we adopt the red-blue color map representation instead of raw event streams to improve efficiency, and uniformly sample 14 frames from each video (or all frames if fewer than 14). 
Similarly, for \textsc{DSEC}, 14 frames are sampled from each segment to maintain training efficiency. 
After annotation, image/video–event–text pairs from \textsc{N-ImageNet}, \textsc{HARDVS}, and \textsc{N-Caltech101} are split into training and validation sets according to categories, while the \textsc{DSEC} training set is used entirely by segmenting scenes into multiple frame groups. 
Categories across splits are kept non-overlapping within each dataset to ensure zero-shot evaluation.

\begin{figure*}
    \centering
    \includegraphics[width=\linewidth]{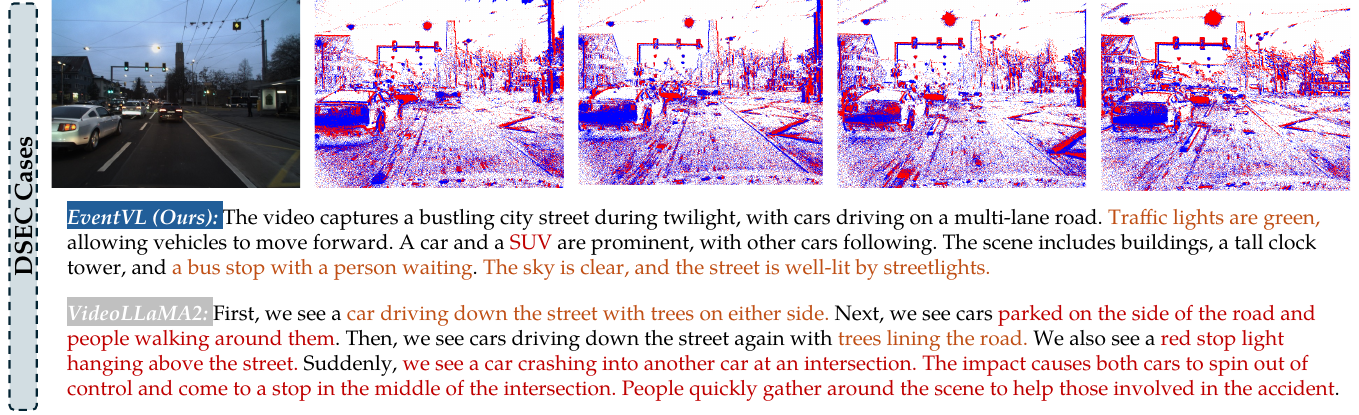}
    \vspace{-0.4cm}
    \caption{Qualitative results about zero-shot caption generation by utilizing \textsc{DSEC}~\cite{Dsec} origin val set from Event-Base dataset. \textcolor{red}{Red color} denotes model wrong description and \textcolor{orange}{orange color} denotes the accurate object description. It shows that \textsc{EventVL} can easily understand the complicated traffic scenes like low light and multiple objects based on event streams.}
    \label{fig:illu2}
    \vspace{-0.3cm}
\end{figure*}

\subsection{Comparison with SOTAs}

We only consider the event data as input for accurately evaluating the model’s event-based understanding ability.

\noindent \textbf{Zero-shot Captioning.} Table~\ref{tab:zsdg} shows that video MLLMs benefit from temporal information, while the lower performance of \textsc{InternVL-E} exposes the event-image gap. \textsc{EventVL} achieves CIDEr scores of 2.619 on \textsc{HARDVS} and 0.599 on N-Caltech, outperforming all baselines with only 2.3B parameters. The improvement indicates that ESR preserves temporal and spatial cues that are lost when event frames are directly passed to an image-based encoder.

\noindent \textbf{Question Answering.} As shown in Table~\ref{tab:vqa}, \textsc{EventVL} obtains the best VQA scores on both \textsc{N-ImageNet} (4.13) and \textsc{HARDVS} (4.26), exceeding adapted MLLM baselines with fewer parameters. This consistent gain across static and dynamic datasets suggests that the learned event-language space supports both factual recognition and event-based reasoning.

\noindent \textbf{Few-shot Evaluation.} Table~\ref{tab:fewshot} shows consistent gains as the number of shots increases, with diminishing improvement at $n\geq20$, consistent with scaling-law behavior~\cite{scalinglaw}.

\begin{table}[!t] 
\centering
\caption{Quantitative results about few-shot caption generation. ``$n$'' denotes the sample number per category used in training.}
\resizebox{0.95\columnwidth}{!}{
\begin{tabular}{lccccc}
\toprule
\textbf{Val set} & $\boldsymbol{n}$ & \textbf{Bleu-4} & \textbf{METEOR} & \textbf{ROUGE-L} & \textbf{CIDEr} \\
\midrule

\multirow{5}{*}{\textsc{N-ImageNet}}  & 1 & 0.058 & 0.103 & 0.308 & 0.398 \\
&  5 & 0.081 & 0.122 & 0.345 & 0.576 \\

& 10 & 0.092 & 0.131 & 0.354 & 0.673 \\
& 15 & 0.094 & 0.133 & 0.365 & 0.699 \\

 & 20 & \textbf{0.099} & \textbf{0.138} & \textbf{0.363} & \textbf{0.742} \\
\midrule

\multirow{5}{*}{\textsc{HARDVS}} & 1 & 0.216 & 0.210 & 0.541 & 1.244 \\
& 5 & 0.343 & 0.278 & 0.607 & 2.017 \\

 & 10 & 0.356 & 0.299 & 0.625 & 2.177 \\
& 15 & 0.378 & 0.314 & 0.628 & 2.457 \\
&  20 & \textbf{0.405} & \textbf{0.318} & \textbf{0.641} & \textbf{2.721} \\
\bottomrule
\end{tabular}
}
\label{tab:fewshot}
\vspace{-0.2cm}
\end{table}

\begin{table}[!t] 
\centering
\caption{Quantitative performance of VQA on different datasets.}
\resizebox{0.9\linewidth}{!}{
\begin{tabular}{l|c|c|c}
\toprule
\textbf{Dataset} & \textbf{Model} & \textbf{\#param} & \textbf{VQA} \\
\midrule

\multirow{4}{*}{\textsc{N-ImageNet}~\cite{N-imagenet}} 
& \cellcolor{gray!20}Video-LLaVA~\cite{Video-llava} & \cellcolor{gray!20}7B & \cellcolor{gray!20}2.87 \\
& \cellcolor{blue!5}\textsc{InternVL-E}~\cite{internvl-2.0} & \cellcolor{blue!5}2.3B & \cellcolor{blue!5}3.34 \\
& \cellcolor{blue!5}\textsc{Qwen3-VL-E} & \cellcolor{blue!5}2.4B & \cellcolor{blue!5}3.86 \\
& \cellcolor{blue!20}\textbf{\textsc{EventVL (Ours)}} & \cellcolor{blue!20}2.3B & \cellcolor{blue!20}\textbf{4.13} \\
\midrule

\multirow{4}{*}{HARDVS~\cite{hardvs}} 
& \cellcolor{gray!20}Video-LLaVA~\cite{Video-llava} & \cellcolor{gray!20}7B & \cellcolor{gray!20}3.23 \\
& \cellcolor{blue!5}\textsc{InternVL-E}~\cite{internvl-2.0} & \cellcolor{blue!5}2.3B & \cellcolor{blue!5}3.73 \\
& \cellcolor{blue!5}\textsc{Qwen3-VL-E} & \cellcolor{blue!5}2.4B & \cellcolor{blue!5}3.95 \\
& \cellcolor{blue!20}\textbf{\textsc{EventVL (Ours)}} & \cellcolor{blue!20}2.3B & \cellcolor{blue!20}\textbf{4.26} \\
\bottomrule

\end{tabular}
}
\vspace{-0.2cm}
\label{tab:vqa}
\end{table}

\subsection{Qualitative results}

Using only events (Fig.~\ref{fig:illu1}), \textsc{EventVL} produces accurate descriptions and recognizes objects, actions, and attributes in difficult scenes, whereas VideoLLaMA2 hallucinates because its training data do not contain event representations. In the HARDVS examples, the model describes human motion and infers attributes such as clothing color from the learned image prior. In driving scenes (Fig.~\ref{fig:illu2}), it gives more precise descriptions with fewer hallucinations, including traffic-light states and object relations that are missed by the video baseline.

\begin{table}[!t]
   \centering
\caption{Ablation study of the proposed Event Spatiotemporal Representation (ESR). ``HTS'' denotes the Hierarchical Temporal Split and ``ASS'' denotes the Adaptive Spatial Split.}
   \resizebox{0.9\columnwidth}{!}{
\begin{tabular}{cc|cccccc}
\toprule
\multirow{2}{*}{HTS} & \multirow{2}{*}{ASS} & \multicolumn{2}{c}{\textbf{\textsc{N-ImageNet}}} & \multicolumn{2}{c}{\textbf{HARDVS}} \\
&   &  ROUGE-L & CIDEr & ROUGE-L & CIDEr \\
\midrule
- & - & 0.352 & 0.806 & 0.493 & 2.274 \\
\checkmark & - & 0.384 & 0.833 & 0.615 & 2.512 \\
- & \checkmark & 0.377 & 0.821 & 0.611 & 2.478 & \\
\checkmark & \checkmark & \textbf{0.397} & \textbf{0.853} & \textbf{0.639} & \textbf{2.619} \\
\bottomrule
\end{tabular}}
\label{tab:HTS/ASS}
\vspace{-0.2cm}
\end{table}

\subsection{Multi-round Dialogue}

\begin{figure}[t]
    \centering
    \vspace{-0.2cm}    \includegraphics[width=\linewidth]{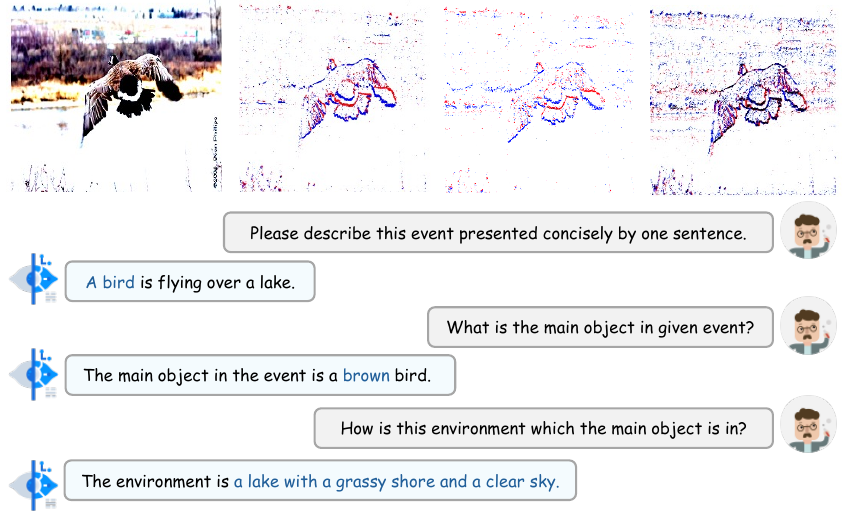}
    \vspace{-0.4cm}
    \caption{Results about multi-round dialogue by utilizing \textsc{N-ImageNet}~\cite{N-imagenet} val set from Event-QA dataset. \textcolor{blue}{Blue color} denotes the detailed description.}
    \label{fig:coversation}
    \vspace{-0.2cm}
\end{figure}

Figure~\ref{fig:coversation} illustrates multi-round dialogue in which \textsc{EventVL} captures scene semantics and object details such as \textcolor{blue}{“a clear sky”}. Across successive questions, the model maintains the event context while refining the answer to the user query, demonstrating that its representation can support interaction beyond single-turn captioning.

\begin{table}[!t]
    \centering
    \caption{Sensitivity analysis of $n_\epsilon$ across static (N-IMAGENET) and high-speed driving (DSEC) scenarios.}
    \label{tab:event_count}
    \resizebox{0.9\columnwidth}{!}{
    \begin{tabular}{c|cc|cc}
    \toprule
    \multirow{2}{*}{$n_\epsilon$} & \multicolumn{2}{c|}{N-IMAGENET (sparse static)} & \multicolumn{2}{c}{DSEC (high-speed driving)} \\
    \cmidrule{2-5}
    & ROUGE-L & CIDEr & ROUGE-L & CIDEr \\
    \midrule
    10,000 & 0.341 & 0.692 & 0.376 & 0.987 \\
    20,000 & 0.378 & 0.794 & 0.412 & 1.184 \\
    \textbf{40,000 (default)} & \textbf{0.397} & \textbf{0.853} & \textbf{0.467} & \textbf{1.392} \\
    60,000 & 0.394 & 0.851 & 0.470 & 1.398 \\
    80,000 & 0.392 & 0.847 & 0.471 & 1.401 \\
    \bottomrule
    \end{tabular}}
\end{table}

\begin{table}[!t]
    \centering
    \caption{Ablation study of ESR against standard multi-scale decomposition baselines under identical settings.}
    \label{tab:esr_vs_standard}
    \resizebox{0.9\columnwidth}{!}{
    \begin{tabular}{c|cc|cc}
    \toprule
    \multirow{2}{*}{Representation} & \multicolumn{2}{c|}{N-IMAGENET} & \multicolumn{2}{c}{HARDVS} \\
    & ROUGE-L & CIDEr & ROUGE-L & CIDEr \\
    \midrule
    Uniform temporal split             & 0.349 & 0.762 & 0.587 & 2.341 \\
    Uniform spatial split (fixed grid) & 0.358 & 0.798 & 0.594 & 2.376 \\
    HTS only                           & 0.384 & 0.833 & 0.615 & 2.512 \\
    ASS only                           & 0.377 & 0.821 & 0.611 & 2.478 \\
    \textbf{ESR (HTS + ASS, Ours)}     & \textbf{0.397} & \textbf{0.853} & \textbf{0.639} & \textbf{2.619} \\
    \bottomrule
    \end{tabular}}
\end{table}

\begin{table}[!t]
   \centering
\caption{Various loss constructions for the training objective in Equation~\ref{opm}. ``Prior'' denotes the image prior and $\mathcal{L}_{c}$ denotes the image--event embedding similarity loss.}
   \resizebox{0.95\columnwidth}{!}{
\begin{tabular}{cccccccc}
\toprule
\multirow{2}{*}{\textbf{Prior}} & \multirow{2}{*}{\textbf{$\mathcal{L}_{c}$}} & \multicolumn{2}{c}{\textbf{\textsc{N-ImageNet}}} & \multicolumn{2}{c}{\textbf{HARDVS}} & \multicolumn{2}{c}{\textbf{\textsc{N-Caltech101}}} \\
   &   & ROUGE-L & CIDEr & ROUGE-L & CIDEr & ROUGE-L & CIDEr \\
\midrule
- & - & 0.321 & 0.740 & 0.534 & 2.036 & 0.330 & 0.475 \\
\checkmark & - & 0.349 & 0.762 & 0.579 & 2.442 & 0.365 & 0.501 \\
- & \checkmark & 0.363 & 0.749 & 0.556 & 2.420 & 0.341 & 0.534 \\
\checkmark & \checkmark & \textbf{0.397} & \textbf{0.853} & \textbf{0.639} & \textbf{2.619} & \textbf{0.477} & \textbf{0.599} \\
\bottomrule
\end{tabular}}
\label{tab:lossfunc}
\end{table}

\begin{table}[!t]
   \centering
\caption{Different $\lambda_1$ and $\lambda_2$ settings for the training objective in Equation~\ref{opm}.}
   \resizebox{\linewidth}{!}{
\begin{tabular}{cccccccc}
\toprule
\multirow{2}{*}{$\lambda_1$} & \multirow{2}{*}{$\lambda_2$} & \multicolumn{2}{c}{\textbf{\textsc{N-ImageNet}}} & \multicolumn{2}{c}{\textbf{HARDVS}} & \multicolumn{2}{c}{\textbf{\textsc{N-Caltech101}}} \\
  &  & ROUGE-L & CIDEr & ROUGE-L & CIDEr & ROUGE-L & CIDEr \\
\midrule
0.2 & 1 & 0.304 & 0.375 & 0.522 & 1.260 & 0.330 & 0.406 \\
0.5 & 1 & 0.325 & 0.449 & 0.564 & 1.580 & 0.337 & 0.444 \\
1 & 0.1 & 0.297 & 0.680 & 0.476 & 1.911 & 0.387 & 0.323 \\
1 & 0.2 & 0.304 & 0.703 & 0.498 & 2.144 & 0.395 & 0.348 \\
1 & 0.5 & 0.352 & 0.739 & 0.585 & 2.308 & 0.421 & 0.460 \\
1 & 1 & \textbf{0.397} & \textbf{0.863} & \textbf{0.639} & \textbf{2.619} & \textbf{0.477} & \textbf{0.599} \\
\bottomrule
\end{tabular}}
\label{tab:lambda}
\vspace{-0.2cm}
\end{table}

\begin{figure}[t]
    \centering
\includegraphics[width=0.8\linewidth]{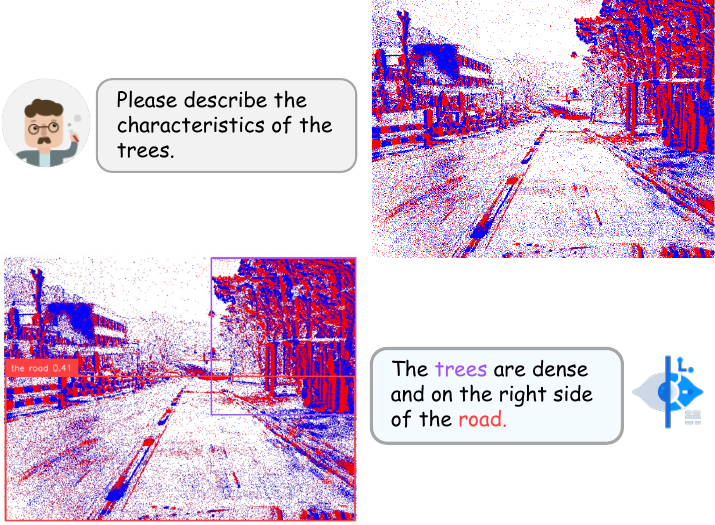}
    \caption{Illustration of the detection by integrating \textsc{EventVL} and GroundingDINO~\cite{Grounding-DINO}. Our \textsc{EventVL} generates a rich description for the specified objects, \textcolor{red}{road} and \textcolor{purple}{trees}, for open-set object detection.}
    \label{fig:det}
\end{figure}

\subsection{Ablation Study}
In this section, we utilize the zero-shot caption generation evaluation setting to perform the comprehensive ablation studies for investigating each effective module. 

\noindent \textbf{Sensitivity of $n_\epsilon$ across diverse scenes.} The event count per frame $n_\epsilon$ is empirically set based on each dataset's average event density (e.g., $\sim$45K on N-IMAGENET, $\sim$22K on N-CALTECH101). To validate its robustness, we conduct sensitivity analyses on N-IMAGENET (sparse static) and a held-out subset of DSEC (high-speed driving). As shown in Table~\ref{tab:event_count}, the two scenarios exhibit complementary patterns: N-IMAGENET shows a single peak at $n_\epsilon = 40K$, where smaller values under-sample motion and larger values introduce padding redundancy; DSEC instead exhibits saturation, where performance plateaus beyond $n_\epsilon = 40K$ as the inherently dense event distribution in high-speed driving streams already provides sufficient signal. Across both regimes, performance degrades gracefully rather than sharply when $n_\epsilon$ deviates from the optimum, indicating that EventVL is robust to $n_\epsilon$ across diverse scene densities, with $n_\epsilon = 40K$ serving as a reliable default.

\begin{table*}[t]
\centering
\begin{minipage}{0.48\textwidth}
    \centering
    \caption{Transferability of ESR across different MLLM backbones. The Qwen3-VL-E variants are trained under single-stage SFT.}
    \label{tab:esr_transfer}
    \resizebox{0.85\linewidth}{!}{
    \begin{tabular}{c|c|cc|cc}
    \toprule
    \multirow{2}{*}{Method} & \multirow{2}{*}{Params} 
    & \multicolumn{2}{c|}{N-IMAGENET} 
    & \multicolumn{2}{c}{HARDVS} \\
    \cmidrule{3-6}
     &  & CIDEr$\uparrow$ & VQA$\uparrow$ & CIDEr$\uparrow$ & VQA$\uparrow$ \\
    \midrule
    Qwen3-VL-E (w/o ESR)        & 2.4B & 0.574 & 3.86 & 2.031 & 3.95 \\
    Qwen3-VL-E + ESR (Ours)     & 2.4B & \textbf{0.713} & \textbf{4.02} & \textbf{2.317} & \textbf{4.11} \\
    \midrule
    EventVL (Ours)              & 2.3B & 0.853 & 4.13 & 2.619 & 4.26 \\
    \bottomrule
    \end{tabular}}
\end{minipage}\hfill
\begin{minipage}{0.48\textwidth}
\centering

\captionof{table}{Different $n_{\max}$ settings for Dynamic Semantic Alignment. Note that we default the $n_{\min}=1$.}
   \resizebox{0.95\columnwidth}{!}{
\begin{tabular}{ccccccc}
\toprule
\multirow{2}{*}{\textbf{$n_{\max}$}} & \multicolumn{2}{c}{\textbf{\textsc{N-ImageNet}}} & \multicolumn{2}{c}{\textbf{HARDVS}} & \multicolumn{2}{c}{\textbf{N-Caltech101}} \\
& ROUGE-L & CIDEr & ROUGE-L & CIDEr & ROUGE-L & CIDEr \\
\midrule
1 & 0.316 & 0.771 & 0.576 & 2.446 & 0.440 & 0.538 \\
5 & 0.348 & 0.826 & 0.615 & 2.626 & 0.468 & 0.580 \\
6 & \textbf{0.397} & 0.853 & 0.639 & 2.619 & 0.477 & 0.599 \\
7 & 0.396 & 0.858 & 0.646 & 2.701 & \textbf{0.482} & \textbf{0.604} \\
8 & 0.394 & \textbf{0.860} & \textbf{0.648} & \textbf{2.702} & 0.477 & 0.586 \\
\bottomrule
\end{tabular}}
\label{tab:dynamicN}
\vspace{-0.2cm}
\end{minipage}
\end{table*}

\begin{table}[!t] 
\centering
\caption{Ablation study of the processed dataset choice for training with event inputs. Categories from training sets do not overlap with test sets. ``w/o im.'' denotes that the image modality is not used.}
\resizebox{0.9\columnwidth}{!}{
\begin{tabular}{l|c|c|cccc}
\toprule
\textbf{Train} & \textbf{Test} & \textbf{Model} & \textbf{Bleu-4} & \textbf{METEOR} & \textbf{ROUGE-L} & \textbf{CIDEr} \\
\midrule

\multirow{6}{*}{+\textsc{N-ImageNet}} & \multirow{2}{*}{N-Caltech101} & - & \textbf{0.445} & \textbf{0.295} & \textbf{0.649} & \textbf{1.302} \\
 &  &  w/o im. & 0.184 & 0.165 & 0.453 & 0.385 \\
\cdashline{2-7}[1pt/1pt]
& \multirow{2}{*}{HARDVS} & - & \textbf{0.187} & \textbf{0.239} & \textbf{0.482} & \textbf{1.010} \\
&  & w/o im. & 0.032 & 0.129 & 0.248 & 0.091 \\
\cdashline{2-7}[1pt/1pt]
& \multirow{2}{*}{\textsc{N-ImageNet}} & - & \textbf{0.212} & \textbf{0.224} & \textbf{0.495} & \textbf{1.789} \\
&  & w/o im. & 0.123 & 0.154 & 0.394 & 0.907 \\
\midrule

\multirow{6}{*}{\makecell[l]{+\textsc{N-ImageNet}\\+\textsc{HARDVS}}} & \multirow{2}{*}{N-Caltech101} & - & \textbf{0.434} & \textbf{0.293} & \textbf{0.647} & \textbf{1.301} \\
 &  & w/o im. & 0.179 & 0.164 & 0.454 & 0.380 \\
\cdashline{2-7}[1pt/1pt]
& \multirow{2}{*}{HARDVS} & - & \textbf{0.445} & \textbf{0.349} & \textbf{0.669} & \textbf{3.015} \\
&  &  w/o im. & 0.394 & 0.323 & 0.630 & 2.572 \\
\cdashline{2-7}[1pt/1pt]
& \multirow{2}{*}{\textsc{N-ImageNet}} & - & \textbf{0.210} & \textbf{0.224} & \textbf{0.495} & \textbf{1.791} \\
&  &  w/o im. & 0.119 & 0.151 & 0.387 & 0.850 \\
\midrule

\multirow{6}{*}{\makecell[l]{+\textsc{N-ImageNet}\\+\textsc{HARDVS}\\+\textsc{DSEC}}} & \multirow{2}{*}{N-Caltech101} & - & \textbf{0.469} & \textbf{0.301} & \textbf{0.651} & \textbf{1.510} \\
 &  & w/o im. & 0.223 & 0.183 & 0.477 & 0.599 \\
\cdashline{2-7}[1pt/1pt]
& \multirow{2}{*}{\textsc{HARDVS}} & - & \textbf{0.445} & \textbf{0.347} & \textbf{0.670} & \textbf{3.055} \\
&  & w/o im. & 0.407 & 0.325 & 0.639 & 2.619 \\
\cdashline{2-7}[1pt/1pt]
& \multirow{2}{*}{\textsc{N-ImageNet}} & - & \textbf{0.210} & \textbf{0.224} & \textbf{0.495} & \textbf{1.805} \\
&  &  w/o im. & 0.125 & 0.157 & 0.397 & 0.853 \\
\bottomrule

\end{tabular}
}
\label{tab:dataset enhanced}
\end{table}


\begin{table}[!t]
   \centering
   \caption{Different level of the event-data aggregation choice for event batch construction. ``L1/L2/L3'' denotes the Level-1/2/3. }
   \resizebox{0.95\columnwidth}{!}{
\begin{tabular}{ccc|cccccc}
\toprule
\multirow{2}{*}{L1} & \multirow{2}{*}{L2} & \multirow{2}{*}{L3} & \multicolumn{2}{c}{\textbf{\textsc{N-ImageNet}}} & \multicolumn{2}{c}{\textbf{HARDVS}} & \multicolumn{2}{c}{\textbf{\textsc{N-Caltech101}}} \\
   &   &   &  ROUGE-L & CIDEr & ROUGE-L & CIDEr & ROUGE-L & CIDEr \\
\midrule
\checkmark & - & - & 0.332 & 0.784 & 0.593 & 1.923 & 0.443 & 0.559 \\
\checkmark & \checkmark & - & 0.356 & 0.805 & 0.610 & 2.457 & 0.454 & 0.565 \\
\checkmark & - & \checkmark & 0.360 & 0.802 & 0.613 & 2.459 & 0.450 & 0.568 \\
\checkmark & \checkmark & \checkmark & \textbf{0.397} & \textbf{0.853} & \textbf{0.639} & \textbf{2.619} & \textbf{0.477} & \textbf{0.599} \\
\bottomrule
\end{tabular}}
\label{tab:level}
\vspace{-0.2cm}
\end{table}

\noindent \textbf{Further Ablation Analysis.} The ablations reveal a consistent pattern across datasets. First, the benefit of ESR is not tied to a particular metric or scene type: combining HTS and ASS improves both ROUGE-L and CIDEr on static N-IMAGENET as well as dynamic HARDVS. The comparison with uniform splits is important because it isolates the design choice from the mere increase in the number of event tokens. Uniform temporal segmentation does not adapt to changes in event density, while a fixed spatial grid can discard thin structures and object boundaries; the adaptive hierarchy preserves both types of information.

The aggregation study provides an additional explanation for this gain. L1 captures local temporal changes, L2 supplies a coarser temporal context, and L3 summarizes the complete stream. Using only one level leaves the decoder with either local motion or global context, which explains the lower CIDEr scores. The full three-level representation is especially beneficial on HARDVS, where actions unfold over longer temporal intervals. Likewise, the $n_{\max}$ study shows that additional patches are useful only until the representation becomes sufficiently fine-grained. Larger values add tokens without a reliable cross-dataset gain, so the default setting balances semantic coverage and computational cost.

\noindent \textbf{Event Representation.} The sensitivity study first shows that the default event count is stable across scene densities: N-IMAGENET peaks at $n_\epsilon=40$K, while DSEC saturates once enough events are retained. This supports using a common operating point without over-tuning to one dataset. In Table~\ref{tab:HTS/ASS}, either HTS or ASS improves over the unstructured representation, but their combination gives the largest gain on both N-IMAGENET and HARDVS. HTS supplies complementary temporal resolutions, whereas ASS preserves high-resolution spatial details; the improvement from their combination therefore indicates that the two modules capture different information rather than duplicating one another. Compared with uniform temporal or spatial splits in Table~\ref{tab:esr_vs_standard}, ESR remains stronger, confirming that hierarchical aggregation and adaptive aspect ratios are important for sparse, nonuniform events.

Table~\ref{tab:dynamicN} shows that a single global patch is insufficient, while increasing $n_{\max}$ improves fine-grained semantics until performance saturates. Although $n_{\max}=7$ or 8 can slightly improve individual metrics, the gain is inconsistent across datasets and increases token and memory cost; we therefore use $n_{\max}=6$ as a balanced setting. Table~\ref{tab:level} further shows that using only one temporal level loses both temporal context and recognition accuracy. Combining L1, L2, and L3 gives the best results, validating the multi-level event batch construction.

\begin{table}[!t]
    \centering
    \caption{Ablation study of DSA against alternative alignment objectives on N-IMAGENET, all integrated with the same ESR backbone.}
    \label{tab:dsa_vs_alt}
    \resizebox{0.6\linewidth}{!}{
    \begin{tabular}{c|cc}
    \toprule
    Alignment Objective & ROUGE-L & CIDEr \\
    \midrule
    No alignment                          & 0.349 & 0.762 \\
    L2 alignment                          & 0.371 & 0.798 \\
    Contrastive alignment                 & 0.366 & 0.785 \\
    \textbf{Cosine alignment (DSA, Ours)} & \textbf{0.397} & \textbf{0.853} \\
    \bottomrule
    \end{tabular}}
\end{table}

\begin{table}[!t]
    \centering
    \caption{Color hallucination analysis on 500 randomly sampled color-related questions from EventVL-QA.}
\label{tab:color_hallucination}
    \resizebox{0.85\linewidth}{!}{
    \begin{tabular}{c|c|ccc}
    \toprule
    Setting & \#Questions & Correct (\%) $\uparrow$ & Hallucinated (\%) $\downarrow$ & Abstain (\%) \\
    \midrule
    Event        & 500 & 53.6 & 16.8 & 29.6 \\
    Event + Image     & 500 & 85.2 & 4.0  & 10.8 \\
    \bottomrule
    \end{tabular}}
\end{table}

\begin{table}[!t]
   \centering
       \caption{Event retrieval results on processed \textsc{N-Caltech101}~\cite{n-caltech}.}
   \resizebox{0.95\columnwidth}{!}{
    \begin{tabular}{ccccccc}
    \toprule
    \multirow{2}{*}{\textbf{Model}} & \multicolumn{3}{c}{\textbf{Text2Event}} & \multicolumn{3}{c}{\textbf{Event2Text}} \\
    &  Recall@1 & Recall@5 & Recall@10 & Recall@1 & Recall@5 & Recall@10 \\
    \midrule
    EventBind~\cite{eventbind} & \textbf{84.2} & \textbf{97.5} & \textbf{100} & \textbf{88.7} & \textbf{98.4} & \textbf{100} \\
    \cellcolor{blue!20}\textbf{\textsc{EventVL (Ours)}} &  \cellcolor{blue!20}80.4 & \cellcolor{blue!20}96.7 & \cellcolor{blue!20}\textbf{100} & \cellcolor{blue!20}76.0 & \cellcolor{blue!20}97.2 & \cellcolor{blue!20}\textbf{100} \\
    \bottomrule
    \end{tabular}}
    \label{tab:retrieval}
\end{table}

\begin{table}[!t]
\centering
\caption{Comparison with EventGPT on VQA benchmarks. Decoding speed is measured in tokens per second on a single NVIDIA RTX 4090 GPU (FP16, batch size = 1).}
\resizebox{0.98\columnwidth}{!}{
\begin{tabular}{l|c|c|c|c}
\toprule
\textbf{Dataset} & \textbf{Model} & \textbf{\#param} & \textbf{VQA} & \textbf{Decoding Speed (tokens/s)} \\
\midrule
\multirow{2}{*}{\textsc{N-ImageNet}~\cite{N-imagenet}} 
& \cellcolor{gray!20}EventGPT (CVPR'24) & \cellcolor{gray!20}7B & \cellcolor{gray!20}3.34 & \cellcolor{gray!20}52 \\
& \cellcolor{blue!20}\textbf{\textsc{EventVL (Ours)}} & \cellcolor{blue!20}2.3B & \cellcolor{blue!20}\textbf{4.13} & \cellcolor{blue!20}\textbf{81} \\
\midrule
\multirow{2}{*}{HARDVS~\cite{hardvs}} 
& \cellcolor{gray!20}EventGPT (CVPR'24) & \cellcolor{gray!20}7B & \cellcolor{gray!20}3.73 & \cellcolor{gray!20}52 \\
& \cellcolor{blue!20}\textbf{\textsc{EventVL (Ours)}} & \cellcolor{blue!20}2.3B & \cellcolor{blue!20}\textbf{4.26} & \cellcolor{blue!20}\textbf{81} \\
\bottomrule
\end{tabular}}
\label{tab:eventgpt_compare}
\end{table}

The optimization ablations support the same complementarity at the semantic level. The image prior supplies attributes that are sparse or absent in events, while $\mathcal{L}_c$ constrains the event and image embeddings to remain locally consistent. Removing either signal weakens generalization, whereas their combination improves all three benchmarks. Cosine alignment is preferable to L2 because the two modalities can have different feature magnitudes, and it is more stable than contrastive alignment when background event patches provide ambiguous negatives. Finally, the training-strategy comparison shows that representation learning and instruction tuning should be separated: encoder warm-up first resolves the event-image gap, after which the language decoder can learn dialogue behavior without destabilizing the event representation.
\noindent \textbf{Optimization Setting.} Table~\ref{tab:lossfunc} isolates the two additional supervision signals. Removing both the image prior and cosine loss gives the weakest results; the image prior provides a clear improvement, while adding $\mathcal{L}_c$ further improves all three datasets. This supports using the image feature space as a semantic anchor rather than relying on event-text supervision alone. Table~\ref{tab:lambda} shows that reducing $\lambda_1$ weakens event-language learning substantially, whereas small changes to $\lambda_2$ mainly affect the balance between alignment and generation. We therefore retain equal weights. Finally, Table~\ref{tab:dsa_vs_alt} shows that cosine alignment is better than both element-wise L2 and contrastive alignment. Directional similarity is better suited to event and image features whose magnitudes differ because of event sparsity and asynchronous sampling.

\noindent \textbf{Two-stage vs. joint training.} Table~\ref{tab:training_strategy} confirms that encoder warm-up followed by instruction tuning is superior to joint training or stage-2-only optimization. The gap is consistent for both CIDEr and VQA, indicating that the first stage establishes a stable event-image representation before the decoder learns task-specific dialogue behavior.

\begin{table}[!t]
    \centering
    \caption{Training strategies under matched data and compute budgets.}
    \label{tab:training_strategy}
    \resizebox{0.7\linewidth}{!}{
    \begin{tabular}{c|cc|cc}
    \toprule
    \multirow{2}{*}{Strategy} & \multicolumn{2}{c|}{N-IMAGENET} & \multicolumn{2}{c}{HARDVS} \\
    \cmidrule{2-5}
    & CIDEr$\uparrow$ & VQA$\uparrow$ & CIDEr$\uparrow$ & VQA$\uparrow$ \\
    \midrule
    Stage-2 only (no encoder warm-up) & 0.612 & 3.58 & 2.087 & 3.71 \\
    Joint training                    & 0.731 & 3.84 & 2.342 & 3.96 \\
    \textbf{Two-stage (Ours)}         & \textbf{0.853} & \textbf{4.13} & \textbf{2.619} & \textbf{4.26} \\
    \bottomrule
    \end{tabular}}
\end{table}

\noindent \textbf{Dataset Quality Assessment.} 
We first evaluate the effectiveness of our proposed dataset. As shown in Table~\ref{tab:dataset performance}, we test an untrained \textsc{EventVL} initialized from \textsc{InternVL2-2B} on all validation sets to assess how well existing SOTA MLLM weights handle the processed event data. After training, our model achieves 1.769 CIDEr on \textsc{N-ImageNet} (+46.3\%) compared to the untrained baseline (1.306 CIDEr). The results show that naive event representations (e.g., red-blue color maps) often confuse MLLMs and degrade multimodal understanding, highlighting the need for specialized event modeling. We further analyze training strategies in Table~\ref{tab:dataset enhanced}. Single-domain training improves performance on its corresponding validation set but generalizes poorly to other datasets. In contrast, multi-domain training significantly improves general event understanding and alleviates overfitting. For example, it yields nearly a 20\% CIDEr improvement on \textsc{N-Caltech101}. In addition, incorporating scene data from \textsc{DSEC} enhances contextual understanding, leading to improved performance on \textsc{HARDVS}. These results demonstrate that our dataset and training strategy effectively improve event-based multimodal understanding.

\noindent \textbf{Color hallucination analysis.} On 500 color-related questions (Table~\ref{tab:color_hallucination}), event-only input yields 53.6\% correctness and 16.8\% hallucination, while adding images raises correctness to 85.2\% and reduces hallucination to 4.0\%.

\noindent \textbf{ESR Transferability.} To validate that ESR is a generalizable plug-and-play module rather than a design specific to EventVL, we integrated ESR into Qwen3-VL-E (a strong MoE-based MLLM baseline) under identical SFT settings, and compared with the vanilla Qwen3-VL-E without ESR. As shown in Table~\ref{tab:esr_transfer}, integrating ESR consistently improves Qwen3-VL-E's performance, with CIDEr gains of +0.139 (24.2\% relative) on N-IMAGENET and +0.286 (14.1\% relative) on HARDVS. While Qwen3-VL-E + ESR narrows the gap toward EventVL, our complete framework still achieves the strongest overall performance, indicating that ESR works synergistically with the other event-specific components (DSA and two-stage progressive training).

\noindent \textbf{Comparison with Concurrent Work.}
We clarify that our work is concurrent with EventGPT \cite{liu2025eventgpt}, as both were released on arXiv during the same period (EventGPT was later accepted to CVPR 2024). Both studies explore extending MLLMs to event-based understanding, representing some of the earliest attempts to adapt generative MLLMs for event streams. As shown in Table~\ref{tab:eventgpt_compare}, despite using only 2.3B parameters (67\% fewer than EventGPT's 7B), EventVL achieves higher VQA scores and faster decoding than EventGPT. These results demonstrate a clear performance--efficiency advantage.

\subsection{Extensive Application}
We evaluate \textsc{EventVL} on retrieval and open-set detection.

\noindent \textbf{Retrieval Tasks.} We evaluate whether the learned event-language space transfers to retrieval. Following~\cite{Llm-adapters}, we freeze the core model and add lightweight attention layers to the encoders plus a projection layer to the decoder, then train with contrastive learning for three epochs. Table~\ref{tab:retrieval} shows performance comparable to EventBind~\cite{eventbind}, confirming that the generative model can support retrieval after parameter-efficient adaptation. The remaining gap is likely caused by the difficulty of using decoder-only LLMs as text encoders~\cite{behnamghader2024llm2vec}.

\noindent \textbf{Detection.} We further test object-level reasoning by coupling \textsc{EventVL} with GroundingDINO~\cite{Grounding-DINO}. Given a query, \textsc{EventVL} first describes candidate objects from the event stream; GroundingDINO then uses the generated description to localize them. As shown in Figure~\ref{fig:det}, the system identifies and localizes \textcolor{purple}{tree} and \textcolor{red}{road} in a low-light scene. This result suggests that event-language reasoning can provide useful semantic proposals for open-set detection, while the final localization remains handled by the detector.

\noindent \textbf{Cross-task analysis.} The two applications test whether the learned event--language space remains useful beyond captioning and VQA. In retrieval, EventVL reaches 100\% Recall@10 in both directions and remains close to EventBind at Recall@5, while the lower Recall@1 primarily reflects the use of a decoder-only language representation rather than a dedicated retrieval encoder. In detection, the generated descriptions preserve object identity and scene relations sufficiently for an external detector to localize queried objects under low illumination. Thus, the experiments indicate that ESR and DSA transfer to both global matching and object-level reasoning; the remaining gaps also suggest clear directions for future work, including a stronger text-side retrieval encoder and quantitative open-set detection benchmarks.

%% file: sec/5_conclusion.tex
\section{Conclusion}

\label{sec:Conclusion}

We introduce \textbf{\textsc{EventVL}}, an event-based MLLM for comprehensive event-stream understanding. By combining an advanced event encoder, Event Spatiotemporal Representation for efficient aggregation, and Dynamic Semantic Alignment for fine-grained feature alignment, \textsc{EventVL} improves event-based understanding and reasoning. We also contribute a high-quality dataset with nearly 1.4 million event-image-text pairs, providing a foundation for future event-based research.
